\documentclass{article}
\usepackage{ijcai25}

\usepackage{times}
\usepackage{soul}
\usepackage{url}
\usepackage[hidelinks]{hyperref}
\usepackage[utf8]{inputenc}
\usepackage[small]{caption}
\usepackage{graphicx}
\usepackage{amsmath}
\usepackage{amsthm}
\usepackage{booktabs}
\usepackage{algorithm}
\usepackage{algorithmic}
\usepackage[switch]{lineno}
\usepackage{listings}  
\usepackage{xcolor}
\usepackage{amsfonts}
\usepackage{amsmath}
\usepackage{multirow}
\usepackage{booktabs}
\usepackage{adjustbox}

\author{
Wenhao Zhu$^1$
\and
Yuhang Xie$^1$
\and
Guojie Song\footnote{Corresponding Author}$^1$\And Xin Zhang$^1$\\
\affiliations
$^{1}$National Key Laboratory of General Artificial Intelligence, School of Intelligence Science and Technology, Peking University\\
\emails
\{wenhaozhu, gjsong, zhang.x\}@pku.edu.cn, yuhangxie@stu.pku.edu.cn
}

\title{EAVIT: Efficient and Accurate Human Value Identification from Text data via LLMs}

\begin{document}

\maketitle

\begin{abstract}
   The rapid evolution of large language models (LLMs) has revolutionized various fields, including the identification and discovery of human values within text data. While traditional NLP models, such as BERT, have been employed for this task, their ability to represent textual data is significantly outperformed by emerging LLMs like GPTs. However, the performance of online LLMs often degrades when handling long contexts required for value identification, which also incurs substantial computational costs. To address these challenges, we propose EAVIT, an efficient and accurate framework for human value identification that combines the strengths of both locally fine-tunable and online black-box LLMs. Our framework employs a value detector—a small, local language model—to generate initial value estimations. These estimations are then used to construct concise input prompts for online LLMs, enabling accurate final value identification. To train the value detector, we introduce explanation-based training and data generation techniques specifically tailored for value identification, alongside sampling strategies to optimize the brevity of LLM input prompts. Our approach effectively reduces the number of input tokens by up to 1/6 compared to directly querying online LLMs, while consistently outperforming traditional NLP methods and other LLM-based strategies.

\end{abstract}
\section{Introduction}
The recent advent of Large Language Models (LLMs), including the Generative Pre-trained Transformer (GPT) models \cite{brown2020language,achiam2023gpt} and Llama \cite{touvron2023Llama}, has marked a pivotal advancement in natural language processing (NLP). One notable application of LLMs is the discovery and identification of human values from text data, such as arguments \cite{kiesel:2022b,kiesel2023semeval}. This task holds significant potential across various domains, supporting value alignment for LLMs \cite{yao2023value}, value-based argument models \cite{atkinson2021value}, and computational psychological studies \cite{alshomary-etal-2022-moral}.

Before the era of LLMs, identifying human values from text data in computational linguistics was typically framed as a multi-label classification problem and addressed using machine learning and NLP models like BERT \cite{devlin2018bert}. However, these models are now being outperformed by modern LLMs in terms of their general text comprehension capabilities. Since most LLMs are accessible only through black-box APIs and fine-tuning them is computationally and economically inefficient, the standard approach for utilizing LLMs in value identification involves constructing tailored prompts for direct queries. For LLMs to effectively identify values from text data, they must first \textit{learn the value system definition}, akin to how humans do. This definition is typically presented as a long context—for instance, the basic Schwartz values definition spans approximately 2.5k tokens \cite{schwartz2012overview}. Including such lengthy definitions in the input prompt is not only costly but has also been shown to degrade performance in context-heavy tasks \cite{liu2023lost}, a finding corroborated by our experiments.


\begin{figure*}[t]
\centering
\includegraphics[width=\linewidth]{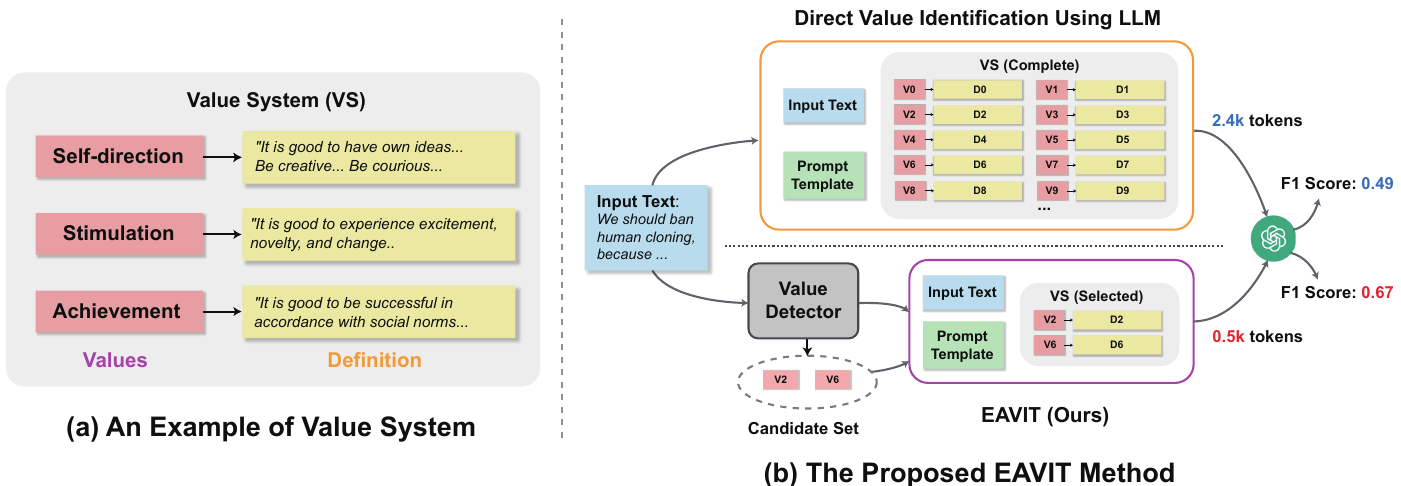}
\caption{An illustration of (a) human value system and (b) the proposed EAVIT method compared with directly using LLMs.}
\label{fig1}
\end{figure*}
Given these challenges, to better leverage the capabilities of LLMs, we propose EAVIT, a framework for \textbf{E}fficient and \textbf{A}ccurate \textbf{H}uman \textbf{V}alue \textbf{I}dentification from Text data. EAVIT begins with a value detector (a tunable local language model) that generates initial value estimations. These estimations are then used to construct a concise input prompt for the LLM, enabling accurate final value identification. The value detector identifies values that are most certainly related and those that are most certainly unrelated, leaving only values with uncertain relevance to be resolved by the LLM. This approach effectively reduces the number of values requiring explicit definition in the input context, thereby shortening the overall context length. To enable the value detector to learn the cognitive logic underlying value identification, we employ an explanation-based fine-tuning method, training the model to reflect on the definitions of values throughout the learning process. To address issues such as insufficient training data and imbalanced class distributions, we draw inspiration from methods like Self-Instruct \cite{wang2022self} and utilize diverse data generation techniques via LLMs to create high-quality training datasets. Additionally, to ensure an optimal candidate set, we perform multiple sampling rounds to identify the most relevant candidate values.

The proposed EAVIT framework not only achieves state-of-the-art performance but also significantly reduces the token cost for inference—down to nearly $\frac{1}{6}$ of the tokens required when directly using LLMs. This makes it a promising and cost-effective solution for large-scale value identification tasks.

Our contributions can be summarized as follows:
\begin{itemize}
\item We introduce EAVIT, a novel framework for identifying human values from text data. This methodology efficiently leverages the power of LLMs, offering accurate and cost-effective value identification results.
\item We employ a diverse set of training strategies for the value detector (local LM in EAVIT), including explanation-based fine-tuning and data generation techniques. Additionally, we develop sampling and prompt-generation methods to create concise and effective input prompts for LLMs.
\item Our approach achieves state-of-the-art performance compared to traditional NLP methods and direct LLM-based strategies. Furthermore, it significantly reduces inference costs to as low as $\frac{1}{6}$ of the tokens required by conventional LLM approaches, making it highly scalable for tasks such as LLM alignment and psychological analysis.
\end{itemize}

\section{Related Works}

\paragraph{Human Value Theories } A detailed review of human value theories in psychology can be found in Appendix. In our paper, following existing works \cite{kiesel2023semeval} in computational linguistics, we adopt Schwartz's Theory of Basic Values \cite{schwartz2012overview} as the basic value system for value identification, which has been applied in multiple fields including economics \cite{ng2005predictors} and LLMs \cite{miotto2022gpt,fischer2023does}. Meanwhile, it should be noted that our method is applicable to any completely defined value system (such as the extended Schwartz value system or those with values set for language models).
\begin{figure}[t]
\centering
\includegraphics[width=0.65\linewidth]{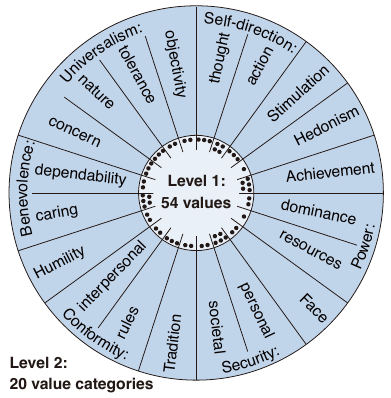}
\caption{An illustration of the Schwartz value systems.}
\label{fig2}
\end{figure}

\paragraph{Value Identification from Text Data.} In NLP, the identification of human values from text data can be perceived as a multi-label classification or regression task described by complex task definition. Recent key related works include \cite{qiu2022valuenet,kiesel:2022b,kiesel2023semeval,ren2024valuebench}. In \cite{qiu2022valuenet}, simple social scenario descriptions from \cite{forbes2020social} were selected and annotated using Schwartz's value taxonomy, establishing the ValueNet dataset for value modeling in language models. \cite{kiesel:2022b} was the first to systematically establish the task of identifying hidden human values from argument data and built a dataset, Webis-ArgValues-22, of 5k size derived from social network data and annotated with Schwartz values by humans. \cite{kiesel2023semeval} extended the work of \cite{kiesel:2022b} with dataset Touché23-ValueEval, expanding the dataset size to 9k and held a public competition at the ACL2023 workshop. \cite{yao2023value} also proposes FULCRA dataset (currently unavailable) that labels LLM outputs to Schwartz human values. Our experiments will use these public, human-annotated datasets as the basis for training and validation. Touché23-ValueEval will be the main dataset.

\section{Human Value Identification - Task and Basic Methods}
\subsection{Task Definition}

We first introduce the formal definition of the human value identification task, generally following \cite{kiesel:2022b,kiesel2023semeval}. For text data $T$, the task of \textbf{human value identification} in this paper is to generate a value label $V_i(T)\in\{0,1\}$ for every human value $V_i$ in a value system $\mathbb V=\{V_1:D_1,\ldots,V_n:D_n\}$, which can be viewed as a multi-label classification task. Each value item $V_i$ has its corresponding definition $D_i$, a paragraph of natural language (see Figure \ref{fig1}) that specifies the meaning of value item. For example, the definition of value \textit{Self-direction: thought} is: \textit{It is good to have own ideas and interests. Contained values and associated arguments of this value: Be creative: arguments towards more creativity or imagination; Be curious ... (omitted)}. The labels are: 0 (the text data has no clear connection to the value item) and 1 (the text data resorts to this value item). Unlike other classification tasks in NLP such as binary sentiment analysis, human value identification focuses on more abstract and complex concepts, requiring deep understanding of both input text and value system definition.

\subsection{Basic LM-based Methods}
\label{sec32}

Naturally, considering that value identification can be viewed as a multi-label classification task, we can employ some straightforward NLP methods and models to address it, including direct fine-tuning and prompt-based methods for LLMs . We will first describe and analyze these simple approaches before introducing EAVIT. 

\paragraph{Fine-tuning}
Local language models that can be fune-tuned without incurring significant costs can be directly trained to fit the task. For encoder models, we can use embedding vector (like [CLS] in BERT \cite{devlin2018bert}) to directly generate results for value identification through a linear layer. For the emerging generative models like GPT-2 \cite{brown2020language} and Llama \cite{touvron2023Llama}, we can no longer extract a clear embedding vector representing the entire input sequence. Instead, we can apply prompt-based supervised fine-tuning \cite{brown2020language}, which involves using prompts that guide the model towards generating outputs that contains the identified values.

\paragraph{Prompt-based Methods}
For black-box models like GPT-4 where fine-tuning is either unavailable or too expensive, we typically can only obtain results in natural language form by prompting the model with input prompt queries. Therefore, an intuitive idea is to first input the complete definition of the value system to LLM for learning, then prompt it to identify values from text data. The main challenge with this approach (we call it \textit{single-step prompting}, which completes the task in 1 LLM API call) is that the performance of LLMs tends to deteriorate with the increase of context length when handling complex tasks defined by context. \cite{liu2023lost} has confirmed, when the context length reaches 2k-4k, the GPT model's ability to understand and remember context information significantly worsens. We also find that in this approach the model tends to point out values that appears in the beginning and end of the definition context, learning to unsatisfactory performance. To reduce the context size, identifying each value component individually is also feasible. But this method increases the total token usage, and result bias becomes more serious (see Section \ref{sec_experiments}).

\section{Method}
In this section we introduce EAVIT which utilizes both local tunable LM and online LLMs. Our method involves three stages: (1) Training value detector; (2) Generating candidate value set; (3) Final value identification using LLMs. Prompt templates can be found in Appendix.

\subsection{Training Value Detector}
Our first objective is to tune a local LM that has the basic value identification capabilities. To achieve this goal, we opt to fine-tune the open-source generative language model Llama2-13b-chat \cite{touvron2023Llama2} as value detector, to equip the base model with robust semantic understanding capabilities. With QLoRA \cite{dettmers2023qlora,hu2021lora}, finetuning Llama2-13b-chat can be executed on 4 Nvidia RTX 4090 GPUs with 24GB VRAM.

\subsubsection{Explanation-based Fine-tuning} 

The process of value identification can be viewed as identifying the semantic association between the text content and values based on their definition. Therefore, correct identification result must be \textit{interpretable}. When training encoder-only models like BERT, we can only use numerical value labels as supervisory signals for fine-tuning. However, when using generative models like Llama, we can include natural language explanations of the value identification process during fine-tuning, which enables the model to gain a much deeper understanding of the task's implications, similar to chain-of-thought \cite{wei2023chainofthought} and \cite{ludan-etal-2023-explanation}.

We use GPT-4o-mini to add explanations to the value identification results in the training dataset. Specifically, for input data $T$ and positively labeled $V_i$, we guide LLM to provide a brief explanation based on the definition of this value item and the input data. For example, for text data \textit{we should ban human cloning as it will only cause issues when you have a bunch of the same humans running around all acting the same} and its corresponding value item \textit{Security: societal}, the explanation could be \textit{The text is related to societal security as it addresses the potential chaos and disorder that could arise from human cloning.} After obtaining explanations for the training data labels, we fine-tune the value detector using the following prompt templates below in Alpaca format \cite{alpaca}. We aim to train the model on the semantic logic behind value identification by forcing it generate explanations for each of its results.

\begin{lstlisting}[frame=single]
// Simplified prompt template
### Instruction:
...
For the following input context, identify relevant values from 20 value items.
Recall the definition of these basic values, then select the values that are most prominently reflected or opposed in the context, and provide your explanation. 
...

### Input:
[INPUT_TEXT]

### Response:
(1) [VALUE_1]. Explanation: [EXPLANATION_1];
...
\end{lstlisting}  
Proved by experiments, this method significantly enhances the reliability and performance.

\subsubsection{Data Generation}

Existing value identification datasets (Webis-ArgValues-22, Touché23-ValueEval) are all manually collected and annotated. While manual annotation enhances data reliability, it also leads to a scarcity of data (Touché23-ValueEval only contains 5.4k training instances). More critically, these real-world datasets present significant label imbalance, as illustrated in Figure \ref{fig_distribution} below. To address the issue, we first follow the \textit{self-instruction} \cite{wang2022self} method by employing in-context learning (ICL) to mimic and generate new data based on the existing annotated data with explanations using GPT-4o-miniGPT-4o-mini, to expand the dataset scale.  In addition to data generation based on the single in-context learning (ICL) method, we also utilize \textit{targeted data generation} to compensate for the distribution imbalances of different value labels in the training dataset. We find the least frequent values and select the corresponding labels that have occurred in the training dataset as target labels, then use them as ICL examples to guide LLM in generating corresponding data. After each round, we filter out the repeated (ROGUE-L similarity $>0.7$) and obvious erroneous data. Additionally, we constantly replace the examples used in in-context learning, ensuring that the value annotations of all examples comprehensively cover the entire value system.

\begin{figure}[h]
\centering
\includegraphics[width=0.95\linewidth]{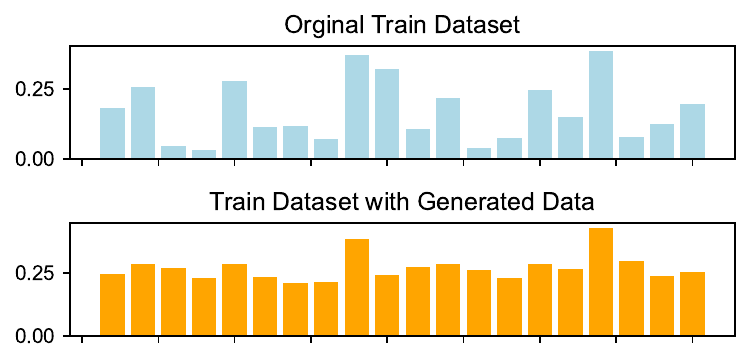}
\caption{Value class distribution of original Touché23-ValueEval train dataset and ours with generated data.}
\label{fig_distribution}
\end{figure}

For Touché23-ValueEval, we have generated approximately 8k ($1.5\times$ original size) instances of data, and significantly compensated the issue of uneven class distribution (see Figure \ref{fig_distribution}). Consistent with existing research \cite{meng2022generating,huang2022large,meng2023tuning}, the generated data is of high quality and effectively enhances the model's generalization capabilities and ability to recognize value labels with lesser distribution.
\subsubsection{Value Definition Reflection}
In addition to previous methods with text-label data, we also want to teach the model the definition of the entire value system through \textit{explicit reflection}. We achieve this goal by forcing the model to reflect on the definition of values during training.

\subsection{Candidate Value Set Generation}
\label{candidate_set}

Next, with a trained value detector that can produce preliminary value identification results, our goal is to obtain a set of candidate values through the preliminary results, which are values \textit{possibly} related to the input text that need LLM for final determination. Our basic assumption is that the value detector will produce outputs that have some random deviations compared to the correct results. We first sample the output $L$ times for each input data $T$ randomly, and calculate the probability of each value being relevant to $T$:
\begin{align*}
    \bar V_i(T)=|\{j:V_i^j(T)=1, j=1,\ldots,L\}|/{T},
\end{align*}
where $V_i^j(T)$ is the label of $V_i$ at $j$-th output. Usually we set $L=5$ to achieve the balance of reducing randomness and efficiency. Next, we set two thresholds $0<p_{\text{low}}<p_{\text{high}}<1$. For all value items $V_i$ with $\bar V_i(T)>p_{\text{high}}$, we directly determine that $V_i(T)$ is related to T, i.e. $V_i(T)=1$; while the value items $V_j$ with positive probability between $p_{\text{low}}$ and $p_{\text{high}}$ constitute our candidate value set $S(T)$, i.e. 
\begin{align*}
    S(T)=\{V_j:p_{\text{low}}\leq\bar V_j(T)\leq p_{\text{high}}\}.
\end{align*}

\begin{table*}[ht]
    \begin{adjustbox}{width=2.1\columnwidth,center}
    \begin{tabular}{l|cccc|cccc|c}
        \toprule
        Dataset & \multicolumn{4}{c|}{Webis-ArgValues-22} & \multicolumn{4}{c|}{Touché23-ValueEval} & \\
        \midrule
        Dataset Split& \multicolumn{2}{c|}{Validation} & \multicolumn{2}{c|}{Test} & \multicolumn{2}{c|}{Validation} & \multicolumn{2}{c|}{Test} & \\
        \midrule
        Metric & Acc & F1-score  & Acc & F1-score & Acc & F1-score & Acc & F1-score &  \#LLM Token \\
        \midrule
        BERT (finetune) & 0.78 & 0.32 & 0.79 & 0.33 & 0.81 & 0.40 & 0.79 & 0.41 & -  \\
        RoBERTa (finetune) & 0.79 & 0.31 & 0.78 & 0.35 & 0.81 & 0.39 & 0.80 & 0.42 & -  \\
        ValueEval'23 Best \cite{schroter2023adam} & - & - & - & - & - & - & - & 0.56 & -  \\
        \midrule
        GPT-2 (finetune+prompting) & 0.75 & 0.34 & 0.76 & 0.33 & 0.72 & 0.30 & 0.73 & 0.34 & -  \\
        Llama2-chat-13b (prompting) & 0.70 & 0.29 & 0.72 & 0.30 & 0.78 & 0.31 & 0.76 & 0.27 & -  \\
        Llama2-chat-13b (finetune+prompting) & 0.86 & 0.42 & 0.84 & 0.44 & 0.82 & 0.41 & 0.82 & 0.45 & -  \\
        \midrule
        GPT-4o-mini (\textit{single-step} prompting)& 0.83 & 0.50 & 0.84 & 0.50 & 0.82 & 0.54 & 0.86 & 0.53 & \multirow{2}{*}{2.4k}   \\
        GPT-4o (\textit{single-step} prompting)& 0.84 & 0.51 & 0.86 & 0.54 & 0.85 & 0.55 & 0.86 & 0.52 &  \\
        \midrule
        GPT-4o-mini (\textit{5-steps} prompting) & 0.83 & 0.49 & 0.82 & 0.50 & 0.84 & 0.49 & 0.85 & 0.52 & \multirow{2}{*}{2.8k}  \\
        GPT-4o (\textit{5-steps} prompting) & 0.86 & 0.51 & 0.84 & 0.50 & 0.87 & 0.50 & 0.87 & 0.54 &  \\
        \midrule
        GPT-4o-mini (\textit{sequential} prompting - simple) & 0.82 & 0.50 & 0.86 & 0.49 & 0.83 & 0.51 & 0.85 & 0.50 & \multirow{2}{*}{3.0k}  \\
        GPT-4o (\textit{sequential} prompting - simple) & 0.82 & 0.52 & 0.84 & 0.51 & 0.87 & 0.55 & 0.89 & 0.54 &  \\
        \midrule
        GPT-4o-mini (\textit{sequential} prompting - CoT) & 0.83 & 0.52 & 0.84 & 0.52 & 0.87 & 0.56 & 0.86 & 0.57 & \multirow{2}{*}{3.6k}  \\
        GPT-4o (\textit{sequential} prompting - CoT) & 0.86 & 0.55 & 0.86 & 0.56 & 0.88 & 0.57 & 0.89 & 0.58 &  \\
        \midrule
        Llama2-chat-13b (EAVIT) & 0.88\small$\pm 0.05$ & 0.52\small$\pm 0.03$ & 0.89\small$\pm 0.07$ & 0.53\small$\pm 0.02$ & 0.88\small$\pm 0.04$ & 0.55\small$\pm 0.03$ & 0.89\small$\pm 0.07$ & 0.57\small$\pm 0.01$ & -  \\
        \midrule
        EAVIT (Llama2-chat-13b + GPT-4o-mini) & 0.93\small$\pm 0.02$ & 0.63\small$\pm 0.04$ & \textbf{0.92}\small$\pm 0.03$ & \textbf{0.63}\small$\pm 0.02$ & \textbf{0.95}\small$\pm 0.01$ & 0.65\small$\pm 0.02$ & \textbf{0.94}\small$\pm 0.02$ & 0.66\small$\pm 0.03$ & \multirow{2}{*}{\textbf{0.45k}}   \\
        EAVIT (Llama2-chat-13b + GPT-4o) & \textbf{0.94}\small$\pm 0.03$ & \textbf{0.65}\small$\pm 0.02$ & \textbf{0.92}\small$\pm 0.02$ & \textbf{0.66}\small$\pm 0.01$ & \textbf{0.95}\small$\pm 0.02$ & \textbf{0.66}\small$\pm 0.01$ & \textbf{0.94}\small$\pm 0.04$ & \textbf{0.69}\small$\pm 0.02$ & \\
        \bottomrule
    \end{tabular}
    \end{adjustbox}
    \caption{Results on Webis-ArgValues-22 and Touché23-ValueEval (level-2 label) dataset.}
    \label{tbl1}
    
\end{table*}

\begin{figure*}[t]
\centering
\includegraphics[width=0.9\linewidth]{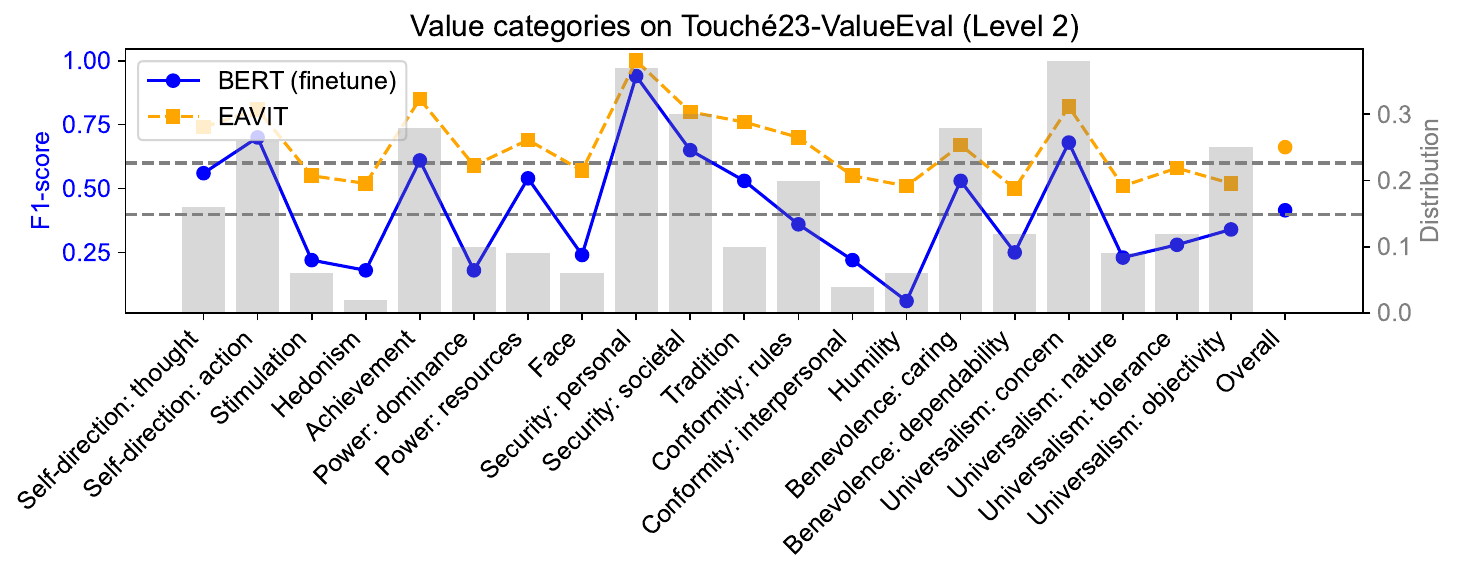}
\caption{Plot of \textbf{F1-scores} and \textbf{class distribution} on the Touché23-ValueEval test set over the labels by level. The grey bars show the label distribution.}
\label{fig_valueeval}
\end{figure*}

\subsection{Final Value Identification via LLMs}
Finally, we use the most acknowledged online LLM, the GPT series including GPT-4, GPT-4o, GPT-4o-mini, to test the values in the candidate set to obtain the final identification results. This is a simple, but most important step in EAVIT. Here, we only need to include the definitions of the values in the candidate set in the LLM prompt. Since the candidate set $S$ is much smaller (3.3 for Touché23-ValueEval) than the entire value system $\mathbb{V}$ (20), the context length of our approach is much shorter than directly using the LLM (Table \ref{tbl1}). More concentrated input prompt can make LLM's output more accurate and reduce the bias caused by forgetting and the order of context content; at the same time, the API cost is significantly lower.

\begin{table*}[ht]
    \begin{adjustbox}{width=1.3\columnwidth,center}
    \begin{tabular}{l|c|c|c}
        \toprule
        Dataset Split& Validation & Test &  \\
        \midrule
        Metric & \multicolumn{2}{c|}{Accuracy} & \#LLM Token / Sample \\
        \midrule
        BERT (Finetune) & 0.60 & 0.61 & -  \\
        RoBERTa (Finetune) & 0.57 & 0.55 & -  \\
        \midrule
        GPT-4o-mini (single-step prompting) & 0.70 & 0.72 & 1.5k  \\
        GPT-4o-mini (sequential prompting) & 0.75 & 0.78 & 1.9k  \\
        \midrule
        EAVIT (Llama2-chat-13b + GPT-4o-mini) & \textbf{0.80} & \textbf{0.78} & \textbf{0.5k}  \\
        
        \bottomrule
    \end{tabular}
    \end{adjustbox}
    \caption{Results on ValueNet (augmented) dataset.}
    \label{tbl2}
\end{table*}

\begin{figure*}[h]
\centering
\includegraphics[width=\linewidth]{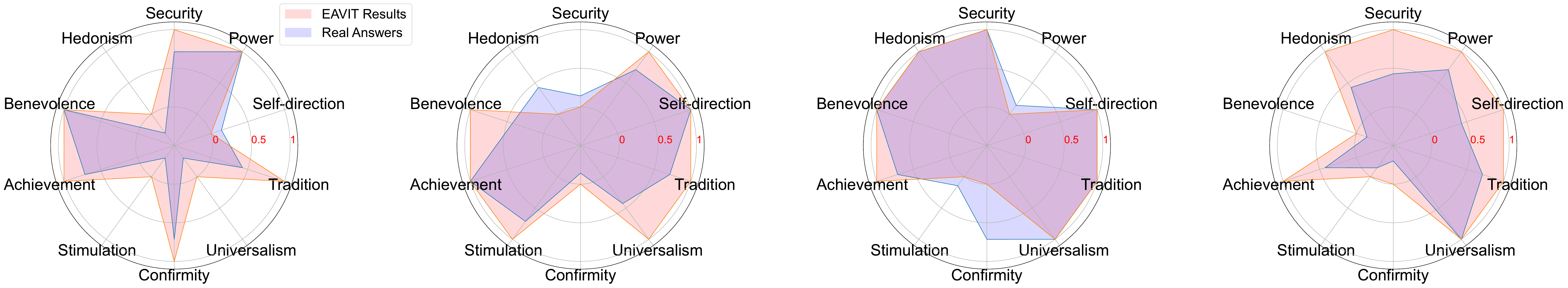}
\caption{Sample Visualization of Value Identification of Virtual Individuals}
\label{fig_casestudy}
\end{figure*}

\section{Experiments}
\label{sec_experiments}
\subsection{Value Identification on Public Datasets}
\subsubsection{Datasets and Methods}

We conducted experiments on three public and manually-labelled datasets: ValueNet (Augmented) \cite{qiu2022valuenet}, Webis-ArgValues-22 \cite{kiesel:2022b}, and Touché23-ValueEval \cite{kiesel2023semeval}. Details can be found in Appendix. In experiments we focus on the Schwartz value systems, but our general method can also be applied to other value systems including \cite{ren2024valuebench} by changing the definitions of the value system and training datasets. For all datasets, we report the accuracy and the \textbf{officially recommended} F1-score on the validation and test data.
We conducted a comparative analysis of our proposed EAVIT approach against various fine-tuning and prompt-based methods. For encoder models BERT \cite{devlin2018bert} and RoBERTa \cite{liu2019roberta}, we directly obtain the prediction results by passing [CLS] token embedding through a linear layer, and train on training dataset. We also reference the  SemEval-2023 Task 4 competition best result \cite{schroter2023adam}, which utilized multiple ensemble RoBERTa models and larger pretrain datasets. For GPT-2 \cite{brown2020language} and Llama2-13b-chat \cite{touvron2023Llama2}, we employ a simple prompt to directly output the value identification results, and report the direct results and fine-tuned results. As we have discussed in Section \ref{sec32}, for online LLMs GPT-4o-mini and GPT-4(o) using OpenAI API \cite{achiam2023gpt}, we adopt \textit{single-step prompting}, \textit{5-steps prompting} and \textit{sequential prompting} with multiple variants. \textit{Single-step prompting} let the model to determine the complete value identification results in a single API call for each data, \textit{sequential prompting} identifies each value individually resulting in multiple LLM API calls (20 for ValueEval'23) for each data, while \textit{5-steps prompting} is a balance of those two methods with 5 API calls per data point. We apply direct prompting and chain-of-thought prompting (\cite{wei2023chainofthought}, CoT) for sequential prompting baselines. For EAVIT, we set $p_\text{low}=0.2, p_\text{high}=0.8$ and report the results of the value detector the entire method. Detailed configurations and prompts can be found in the appendix. We report the average and std of 3 random individual runs.

\subsubsection{Results}
Table \ref{tbl1},\ref{tbl2} and Figure \ref{fig_valueeval} show the experiment results. 

\paragraph{Performance and efficiency of EAVIT.} We can observe that EAVIT significantly improves accuracy while only using up to 1/7 LLM tokens compared to directly using LLMs. Comparing with directly fine-tuning, EAVIT's performance on values that appear less frequently on training data is noticeably better, demonstrating the effectiveness of data generation and explain-based fine-tuning. Furthermore, the performance of the EAVIT method surpasses that of standalone local LMs or online LLM methods, indicating that their combination can effectively complement the shortcomings - the local LM's relatively weak textual understanding power and the substantial impact of context on performance and efficiency of online LLMs - to achieve a more optimal balance between performance and efficiency.

\paragraph{Prompting Methods.} The context length in single-step prompting is typically longer, which generally results in lower accuracy than models that have been fine-tuned for aligned models. However, when using sequential prompting, even though the query context is more precise, the overall accuracy does not make much difference. According to our observations, this is probably because LLMs have a tendency to provide affirmative responses ($>60\%$ in our experiments) due to alignment \cite{perez2022discovering}. By using Chain-of-Thought, this problem is partly alleviated (reduced to 40\%) and the performance is significantly improved.  

\paragraph{GPT-4o-mini v.s. GPT-4o.} In most experiments, the performances of GPT-4o and its mini version are similar. Considering the API cost, we suggest that GPT-4o-mini is sufficient in most practical human value identification tasks.

Additional studies of API cost and candidate value set thresholds are also provided in Appendix \ref{asec1} and \ref{asec2}.

\subsection{Case Study: Value Identification of Virtual Individuals}
Next, we conduct a hypothetical experiment to simulate the process of human value identification of individuals. The objective is to compare the uniformity and differences between the individual values measured by traditional \textit{psychological questionnaires} and our \textit{text-based} value identification method. We use the public real-human questionnaire results from the World Values Survey (WVS) wave 6 \cite{inglehart2000world,wvsv6}, which is an extensive project that has been conducting value tests on populations worldwide for decades. We selected the responses of 20 real individuals to 10 questions about the Schwartz value system. These questions (v70-v79) sequentially correspond to 10 Schwartz values, and can be considered as a standard value questionnaire in psychology. For each individual, we input the content of these questions and their responses to these questions into GPT-4, and guided GPT-4 to mimic an \textit{virtual individual possessing these values} through prompts \cite{aher2023using}. Subsequently, we selected 20 social topics and guided the simulated individuals to express and explain their views on these social topics, forming 20 text data instances for each individual with format similar to Touché23-ValueEval. Finally, we process these text data using EAVIT and other methods trained on Touché23-ValueEval to identify values behind them, aggregate the results of each virtual individual's text data, and compare them with psychological questionnaire answers. The experiment details are provided in the appendix. 

\begin{table}[t]
    \begin{adjustbox}{width=0.9\columnwidth,center}
    \begin{tabular}{l|c}
        \toprule
        Method & Mean Accuracy \\
        \midrule
        RoBERTa (Finetune) & 0.62 \\
        \midrule
        GPT-4o (single-step prompting) & 0.71  \\
        GPT-4o (sequential prompting) & 0.65  \\
        \midrule
        EAVIT & \textbf{0.78}  \\
        \bottomrule
    \end{tabular}
    \end{adjustbox}
    \caption{Results of virtual individual value identification. EAVIT uses Llama2-chat-13b + GPT-4o-mini here.}
    \label{tbl_casestudy}
\end{table}

\paragraph{Results} Our results and visualizations are presented in Table \ref{tbl_casestudy} and Figure \ref{fig_casestudy}. Our findings indicate that by conducting value identification on the viewpoints data generated by virtual individuals, we can effectively infer their value presets, with a high level of consistency with psychological questionnaires based on values. Similarly, it can be observed from Table \ref{tbl_casestudy} that the performance of EAVIT on this task outperforms both simple LMs and naive usage of LLMs. This experiment uncovers an intriguing potential: we can use large models to conduct value identification on passively collected text data generated by individuals, providing a measure of an individual's values without resorting to the active collection methods in psychology like questionnaires \cite{schwartz2001extending}. This approach, based on LLMs and passively collected data, has the advantages of low data collection cost, high credibility, and strong non-falsifiability. For example, a selfish person is unlikely to answer "no" to the questionnaire item "Do you care about others?", but its behavior could likely be reflected in social network traces.

\subsection{Training Value Detector: Ablation Study}
\begin{figure}[h]
\centering
\includegraphics[width=0.9\linewidth]{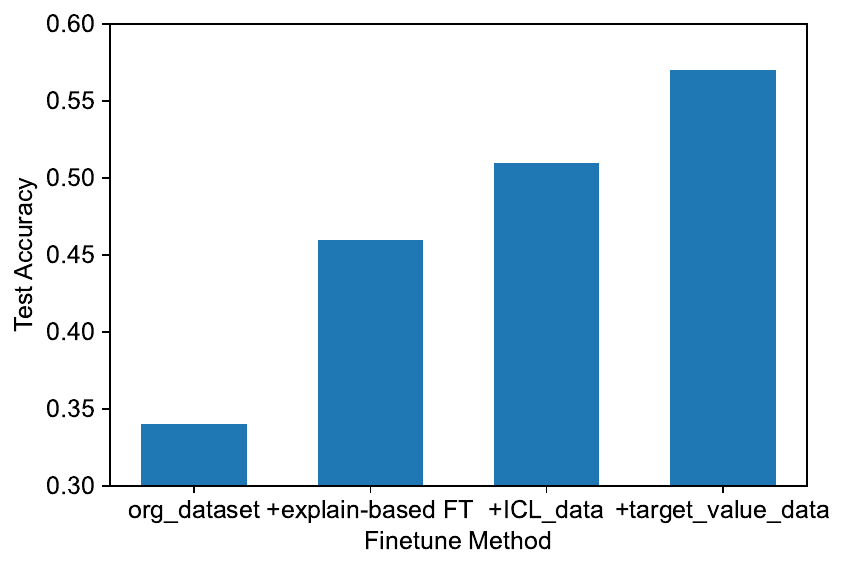}
\caption{Results of different models with different finetune strategies.}
\label{fig_ft}
\end{figure}

We now investigate the impact of fine-tuning strategies and data generation on the performance of the value detector model on Touché23-ValueEval dataset. In order to directly investigate the impact of different fine-tuning methods and data generation techniques, we train four different versions of the Llama2-13b-chat value detector models:
\textit{+org\_dataset}: model fine-tuned on the original dataset using direct prompts;
\textit{+explain-based FT}: model with explanation-based fine-tuning on the original dataset.
\textit{+icl\_data}: the previous model with an additional 4k training data entries generated using a simple ICL;
\textit{+target\_value\_data}: the previous model with an additional 4k training data entries specifically generated for less frequent target values.
We then plot the performance of these models in Figure \ref{fig_ft}.
The results demonstrate that the explain-based finetune method and data generation effectively enhances the model's performance.

\subsection{Output Consistency}
\begin{figure}[h]
\centering
\includegraphics[width=0.9\linewidth]{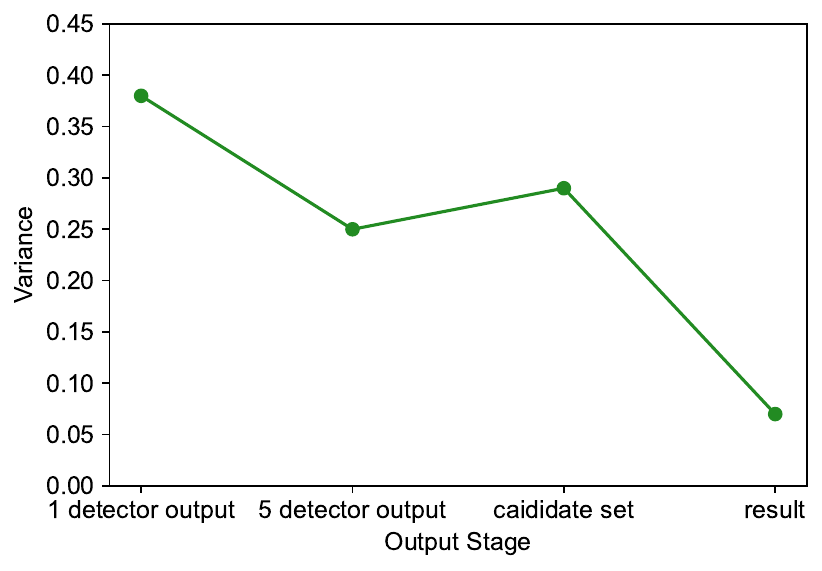}
\caption{Output variance at different output stages.}
\label{fig_var}
\end{figure}

To ensure the feasibility of our method in practical applications such as large-scale data analysis, it is crucial to achieve high output consistency. To this end, we randomly extracted 200 data points from the Touché23-ValueEval test dataset. For each data point, we randomly sampled the results 10 times at different output stages (\textit{1 detector output}: single output from the value detector; \textit{5 detector output}: average of 5 outputs from the value detector, \textit{candidate set}: see Section \ref{candidate_set}, and \textit{final result}). We computed the average variance of 10 samples, with each output viewed as a 20-dimension vector. The results are presented in Figure\ref{fig_var}.

We observe that the single output from the value detector exhibited high randomness, but sampling effectively reduces it. Despite the increased randomness in the candidate set (which indicates that different outputs from the value detector may favor different random values), the final decision made by the LLMs effectively eliminated irrelevant random errors, achieving stable, high-quality results. This suggests that our method could be applied to scenarios that require high output stability.
\section{Conclusion}

This paper investigates the potential of LLMs for identifying human values from text data. Despite challenges, LLMs have demonstrated superior capabilities compared to previous methods. Our work of efficient and accurate value identification can have potential usage for value alignment in LLMs, social network analysis, and psychological studies.
\section{Ethical Considerations}
Our experiments only utilize existing public models and datasets (may include human-generated or human-labeled data) available for research, currently there are no ethical issues. However, our method involves inferring and measuring human values, and potential ethical risks must be carefully considered in possible practical applications.

\clearpage

\section*{Acknowledgements}
This work was supported by the National Natural Science Foundation of China (Grant No. 62276006).

\bibliographystyle{named}
\bibliography{ijcai25}

\clearpage
\section{Appendix}
\label{sec:appendix}
\subsection{Prompt Templates}
\subsubsection{Prompt Template for Generating Explanation}

\begin{lstlisting}[frame=single]
// Simplified prompt template
...
You are an expert on Schwartz Theory of Basic Values.
Below is the definition of human value [VALUE]:
[DEFINITION]


For the following text, can you explain how the text is related to the human value [VALUE]?
[INPUT_CONTEXT]
Your answer should be as concise as possible, generate briefly and clear explanations.
Your answer should be less than 20 tokens.
\end{lstlisting}  

\subsubsection{Prompt Template for Transforming ValueEval Data into Plain Text}

\begin{lstlisting}[frame=single]
    I am {data['Stance']} the opinion of {data['Conclusion']}, because {data['Premise']}.
\end{lstlisting}
\subsubsection{Prompt Template for Explanation-based Finetuning}

\begin{lstlisting}[frame=single]
// Simplified prompt template
...
Below is an instruction that describes a task, paired with an input that provides further context. Write a response that appropriately completes the request.

### Instruction:
This is an annotation task to identify and categorize the values based on Schwartz Theory of Basic Values.
For the following input context, identify relevant values from 20 value items.
The 20 basic values are: 'Self-direction: thought', 'Self-direction: action', 'Stimulation', 'Hedonism', 'Achievement', 'Power: dominance', 'Power: resources', 'Face', 'Security: personal', 'Security: societal', 'Tradition', 'Conformity: rules', 'Conformity: interpersonal', 'Humility', 'Benevolence: caring', 'Benevolence: dependability', 'Universalism: concern', 'Universalism: nature', 'Universalism: tolerance', 'Universalism: objectivity'.
Recall the definition of these basic values, then select the values that are most prominently reflected or opposed in the context, and provide your explanation. 
If a value has no apparent connection with input context, do not include it in the final result list.
You should not generate repeated values in the final result list. Your answer should be as short as possible.

### Input:
[INPUT_DATA]

### Response:
(1) [VALUE_1]. Explanation: [EXPLAINATION_1];
(2) [VALUE_2]. Explanation: [EXPLAINATION_2];
...
\end{lstlisting}

\subsubsection{Prompt Template for Value Definition Reflection}
\begin{lstlisting}[frame=single]
// Simplified prompt template
...
Below is an instruction that describes a task, paired with an input that provides further context. Write a response that appropriately completes the request.

### Instruction:
This is a preliminary task to reflect on the definition of Schwartz Theory of Basic Values.
Answer the following question based on your understanding of the theory.

### Input:
What is the definition of [VALUE] in Schwartz Theory of Basic Values?

### Response:
[DEFINITION]
\end{lstlisting}

\subsubsection{Prompt Template for Simple ICL Data Generation}
\begin{lstlisting}[frame=single]
// Simplified prompt template
...
Based on the following human value identification data with labels and explanations, try to come up with more data instances. Try to generate multiple examples when possible.

DATA 1: [INPUT_TEXT_1] - [VALUE_LABELS_EXPLANATIONS_1]
DATA 2: [INPUT_TEXT_2] - [VALUE_LABELS_EXPLANATIONS_2]
...
DATA 8: [INPUT_TEXT_8] - [VALUE_LABELS_EXPLANATIONS_8]
DATA 9:
\end{lstlisting}  
\subsubsection{Prompt Template for Targeted Data Generation}
\begin{lstlisting}[frame=single]
// Simplified prompt template
...
Next, your input will be the definition of some human values in Schwartz value system.
Can you generate text that resorts to these values? 
...
Example:
Input: Self-direction: thought, Definition: It is good...
        Hedonism, Definition: It is good to experience...
Output: Text - We should encourage...
         Explanation: This text resorts to...

Input: [TARGET_VALUE_1], [TARGET_DEFINITION_1],
        [TARGET_VALUE_2], [TARGET_DEFINITION_2],
Output:
\end{lstlisting}  

\subsubsection{Prompt Template for Value Identification via LLMs}

This prompt template is used for final value identification in EAVIT and baseline methods for GPT models. For single-step prompting the full definition of target value system is integrated into the template, while for sequential prompting we ask each value item individually, and for 5-steps prompting we ask 4 value items every input.
\begin{lstlisting}[frame=single]
// Simplified prompt template
...

You are an expert on Schwartz Theory of Basic Values.

### Instruction:
This is an annotation task to identify and categorize the values based on Schwartz Theory of Basic Values.
For the following input context and given human values with definition, identify relevant values from the given value items.
For each value, identify it as 'Relevant' or 'Irrelevant', and justify your answer based on the context and value definition.

##### Example:
(ICL Example)
### Input:
Human values:  Self-direction: thought, Definition: It is good...
        Hedonism, Definition: It is good to experience..

Text: We should encourage...

### Output:
Self-direction: thought - Relevant. Explanation: This text resorts to self-direction: thought because...
Hedonism - Irrelevant. Explanation: This text does not relates to self-direction: thought because...

##### End of Example

### Input:
Human values:  [TARGET_VALUE_1], [TARGET_DEFINITION_1],
                [TARGET_VALUE_2], [TARGET_DEFINITION_2],
Text: [INPUT_TEXT]

### Output:
\end{lstlisting}  

\subsubsection{Prompt Template for Chain-of-Thought Prompting for Value Identification via LLMs}

This prompt template is used for Chain-of-Thought value identification for baseline GPT models (\textit{sequential} prompting - CoT). We introduces a Chain-of-Thought process that encourages the LLM to first discover the relations between input text and value definition, then generate the label result.

\begin{lstlisting}[frame=single]
// Simplified prompt template
...

You are an expert on Schwartz Theory of Basic Values.

### Instruction:
This is an annotation task to identify and categorize the values based on Schwartz Theory of Basic Values.
For the following input context and given human value with definition, identify whether the input human value resorts to the text or not.
Think step by step, first identify certain parts in the input text that are associated with the value, then identify it as 'Relevant' or 'Irrelevant'.

##### Example:
### Input:
Human value:  Self-direction: thought, Definition: It is good to ...

Text: We should encourage students to...

### Output:
First, the text encorages creativity and freedom of thought in school classes. This is associated to "Be creative" and "Have freedom of thought" in the definition of Self-direction: thought.

As a result, I identify Self-direction: thought as 'Relevant'.

##### End of Example

### Input:
Human values:  [TARGET_VALUE_1], [TARGET_DEFINITION_1],
                ...
Text: [INPUT_TEXT]

### Output:
\end{lstlisting}

\subsubsection{Value Definition of Schwartz Value System (Touché23-ValueEval)}

\begin{lstlisting}
    {
    "Self-direction: thought": [
        "It is good to have own ideas and interests. Contained values and associated arguments (examples): Be creative: arguments towards more creativity or imagination; Be curious: arguments towards more curiosity, discoveries, or general interestingness; Have freedom of thought: arguments toward people figuring things out on their own, towards less censorship, or towards less influence on thoughts."
    ],
    "Self-direction: action": [
        "It is good to determine one's own actions. Contained values and associated arguments (examples): Be choosing own goals: arguments towards allowing people to choose what is best for them, to decide on their life, and to follow their dreams; Be independent: arguments towards allowing people to plan on their own and not ask for consent; Have freedom of action: arguments towards allowing people to be self-determined and able to do what they want; Have privacy: arguments towards allowing for private spaces, time alone, and less surveillance, or towards more control on what to disclose and to whom."
    ],
    "Stimulation": [
        "It is good to experience excitement, novelty, and change. Contained values and associated arguments (examples): Have an exciting life: arguments towards allowing people to experience foreign places and special activities or having perspective-changing experiences; Have a varied life: arguments towards allowing people to engage in many activities and change parts of their life or towards promoting local clubs (sports, ...); Be daring: arguments towards more risk-taking."
    ],
    "Hedonism": [
        "It is good to experience pleasure and sensual gratification. Contained values and associated arguments (examples): Have pleasure: arguments towards making life enjoyable or providing leisure, opportunities to have fun, and sensual gratification."
    ],
    "Achievement": [
        "It is good to be successful in accordance with social norms. Contained values and associated arguments (examples): Be ambitious: arguments towards allowing for ambitions and climbing up the social ladder; Have success: arguments towards allowing for success and recognizing achievements; Be capable: arguments towards acquiring competence in certain tasks, being more effective, and showing competence in solving tasks; Be intellectual: arguments towards acquiring high cognitive skills, being more reflective, and showing intelligence; Be courageous: arguments towards being more courageous and having people stand up for their beliefs."
    ],
    "Power: dominance": [
        "It is good to be in positions of control over others. Contained values and associated arguments (examples): Have influence: arguments towards having more people to ask for a favor, more influence, and more ways to control events; Have the right to command: arguments towards allowing the right people to take command, putting experts in charge, and clearer hierarchies of command, or towards fostering leadership."
    ],
    "Power: resources": [
        "It is good to have material possessions and social resources. Contained values and associated arguments (examples): Have wealth: arguments towards allowing people to gain wealth and material possession, show their wealth, and exercise control through wealth, or towards financial prosperity."
    ],
    "Face": [
        "It is good to maintain one's public image. Contained values and associated arguments (examples): Have social recognition: arguments towards allowing people to gain respect and social recognition or avoid humiliation; Have a good reputation: arguments towards allowing people to build up their reputation, protect their public image, and spread reputation."
    ],
    "Security: personal": [
        "It is good to have a secure immediate environment. Contained values and associated arguments (examples): Have a sense of belonging: arguments towards allowing people to establish, join, and stay in groups, show their group membership, and show that they care for each other, or towards fostering a sense of belonging; Have good health: arguments towards avoiding diseases, preserving health, or having physiological and mental well-being; Have no debts: arguments towards avoiding indebtedness and having people return favors; Be neat and tidy: arguments towards being more clean, neat, or orderly; Have a comfortable life: arguments towards providing subsistence income, having no financial worries, and having a prosperous life, or towards resulting in a higher general happiness."
    ],
    "Security: societal": [
        "It is good to have a secure and stable wider society. Contained values and associated arguments (examples): Have a safe country: arguments towards a state that can better act on crimes, and defend or care for its citizens, or towards a stronger state in general; Have a stable society: arguments towards accepting or maintaining the existing social structure or towards preventing chaos and disorder at a societal level."
    ],
    "Tradition": [
        "It is good to maintain cultural, family, or religious traditions. Contained values and associated arguments (examples): Be respecting traditions: arguments towards allowing to follow one's family's customs, honoring traditional practices, maintaining traditional values and ways of thinking, or promoting the preservation of customs; Be holding religious faith: arguments towards allowing the customs of a religion and to devote one's life to their faith, or towards promoting piety and the spreading of one's religion."
    ],
    "Conformity: rules": [
        "It is good to comply with rules, laws, and formal obligations. Contained values and associated arguments (examples): Be compliant: arguments towards abiding to laws or rules and promoting to meet one's obligations or recognizing people who do; Be self-disciplined: arguments towards fostering to exercise restraint, follow rules even when no-one is watching, and to set rules for oneself; Be behaving properly: arguments towards avoiding to violate informal rules or social conventions or towards fostering good manners."
    ],
    "Conformity: interpersonal": [
        "It is good to avoid upsetting or harming others. Contained values and associated arguments (examples): Be polite: arguments towards avoiding to upset other people, taking others into account, and being less annoying for others; Be honoring elders: arguments towards following one's parents or showing faith and respect towards one's elders."
    ],
    "Humility": [
        "It is good to recognize one's own insignificance in the larger scheme of things. Contained values and associated arguments (examples): Be humble: arguments towards demoting arrogance, bragging, and thinking one deserves more than other people, or towards emphasizing the successful group over single persons and giving back to society; Have life accepted as is: arguments towards accepting one's fate, submitting to life's circumstances, and being satisfied with what one has."
    ],
    "Benevolence: caring": [
        "It is good to work for the welfare of one's group's members. Contained values and associated arguments (examples): Be helpful: arguments towards helping the people in one's group and promoting to work for the welfare of others in one group. Be honest: arguments towards being more honest and recognizing people for their honesty; Be forgiving: arguments towards allowing people to forgive each other, giving people a second chance, and being merciful, or towards providing paths to redemption; Have the own family secured: arguments towards allowing people to have, protect, and care for their family; Be loving: arguments towards fostering close relationships and placing the well-being of others above the own, or towards allowing to show affection, compassion, and sympathy."
    ],
    "Benevolence: dependability": [
        "It is good to be a reliable and trustworthy member of one's group. Contained values and associated arguments (examples): Be responsible: arguments towards clear responsibilities, fostering confidence, and promoting reliability; Have loyalty towards friends: arguments towards being a dependable, trustworthy, and loyal friend, or towards allowing to give friends a full backing."
    ],
    "Universalism: concern": [
        "It is good to strive for equality, justice, and protection for all people. Contained values and associated arguments (examples): Have equality: arguments towards fostering people of a lower social status, helping poorer regions of the world, providing all people with equal opportunities in life, and resulting in a world were success is less determined by birth; Be just: arguments towards allowing justice to be 'blind' to irrelevant aspects of a case, promoting fairness in competitions, protecting the weak and vulnerable in society, and resulting a world were people are less discriminated based on race, gender, and so on, or towards fostering a general sense for justice; Have a world at peace: arguments towards nations ceasing fire, avoiding conflicts, and ending wars, or promoting to see peace as fragile and precious or to care for all of humanity."
    ],
    "Universalism: nature": [
        "It is good to preserve the natural environment. Contained values and associated arguments (examples): Be protecting the environment: arguments towards avoiding pollution, fostering to care for nature, or promoting programs to restore nature; Have harmony with nature: arguments towards avoiding chemicals and genetically modified organisms (especially in nutrition), or towards treating animals and plants like having souls, promoting a life in harmony with nature, and resulting in more people reflecting the consequences of their actions towards the environment; Have a world of beauty: arguments towards allowing people to experience art and stand in awe of nature, or towards promoting the beauty of nature and the fine arts."
    ],
    "Universalism: tolerance": [
        "It is good to accept and try to understand those who are different from oneself. Contained values and associated arguments (examples): Be broadminded: arguments towards allowing for discussion between groups, clearing up with prejudices, listening to people who are different from oneself, and promoting to life within a different group for some time, or towards promoting tolerance between all kinds of people and groups in general; Have the wisdom to accept others: arguments towards allowing people to accept disagreements and people even when one disagrees with them, to promote a mature understanding of different opinions, or to decrease partisanship or fanaticism."
    ],
    "Universalism: objectivity": [
        "It is good to search for the truth and think in a rational and unbiased way. Contained values and associated arguments (examples): Be logical: arguments towards going for the numbers instead of gut feeling, towards a rational, focused, and consistent way of thinking, towards a rational analysis of circumstances, or towards promoting the scientific method; Have an objective view: arguments towards fostering to seek the truth, to take on a neutral perspective, to form an unbiased opinion, and to weigh all pros and cons, or towards providing people with the means to make informed decisions."
    ]
}
\end{lstlisting}

\subsection{Experiment Details}
\subsubsection{Value Identification on Public Datasets}

\paragraph{Datasets.}We conducted experiments on three public and manually-labelled datasets: ValueNet (Augmented) \cite{qiu2022valuenet}, Webis-ArgValues-22 \cite{kiesel:2022b}, and Touché23-ValueEval \cite{kiesel2023semeval}. The ValueNet dataset contains 21,374 simple texts describing human behavior, each related to a value in the target value system, comprising a total of 10 Schwartz values. The Webis-ArgValues-22 dataset contains 5,270 real opinion-argument data obtained from internet platforms worldwide, annotated for values using crowdsourcing methods. Its target value system consists of 20 level-2 Schwartz values. Touché23-ValueEval is an improved version of Webis-ArgValues-22, with an expanded data scale of 9,324 entries. The data acquisition method and annotation quality have been improved, and a public competition was held at the ACL workshop.

\paragraph{Method.}We conducted a comparative analysis of our proposed EAVIT approach against various fine-tuning and prompt-based methods. For encoder models BERT \cite{devlin2018bert} and RoBERTa \cite{liu2019roberta}, we directly obtain the prediction results by passing [CLS] token embedding through a linear layer, and train on training dataset. We also reference the best results from the  SemEval-2023 Task 4 competition \cite{schroter2023adam}, which utilized multiple ensemble RoBERTa models and pretrained them on a larger scale corpus dataset. For language models GPT-2 \cite{brown2020language} and Llama2-13b-chat \cite{touvron2023Llama2}, we employ a simple prompt to directly output the value identification results, and we record the results of directly using this prompt, as well as the results after fine-tuning with this prompt. For online LLMs we adopt \textit{single-step prompting} and \textit{sequential prompting} as discussed. \textit{Single-step prompting} let the model to determine the complete value identification results in a single API call for each data. Meanwhile, \textit{sequential prompting} identifies each value individually, resulting in multiple LLM API calls (20 for ValueEval'23) for each data, reducing the context size of every single query but increasing the total token usage. For EAVIT, we report the results of the separate value detector (average of 5 output samples) and the results of the entire method. The base model of value detector is a Llama2-13b-chat \cite{touvron2023Llama2} model finetuned with QLoRA \cite{dettmers2023qlora,hu2021lora} and 4-bit quantinization, and the training takes up to 10 hours on a Nvidia RTX 4090 GPU.

\subsubsection{Case Study: Value Identification of Virtual Individuals}

We use the real-human questionnaire results from the World Values Survey (WVS) wave 6 \cite{inglehart2000world,wvsv6}, which is an extensive project that has been conducting value tests on populations worldwide for decades. We selected the responses of 20 real individuals to 10 questions about the Schwartz value system. This set of questions are:
\begin{lstlisting}

Now I will briefly describe some people. Using this card, would you please indicate for each description whether that person is very much like you, like you, somewhat like you, not like you, or not at all like you?
(Code one answer for each description): 

V70. It is important to this person to think up new ideas and be creative; to do things one's own way.
V71. It is important to this person to be rich; to have a lot of money and expensive things.
V72. Living in secure surroundings is important to this person; to avoid anything that might be dangerous.
V73. It is important to this person to have a good time; to spoil oneself.
V74. It is important to this person to do something for the good of society.
V75. Being very successful is important to this person; to have people recognize one's achievements.
V76. Adventure and taking risks are important to this person; to have an exciting life.
V77. It is important to this person to always behave properly; to avoid doing anything people would say is wrong.
V78. Looking after the environment is important to this person; to care for nature and save life resources.
V79. Tradition is important to this person; to follow the customs handed down by one's religion or family.
\end{lstlisting}

The answers can be selected from:
\begin{lstlisting}
Very much like me (1)
Like me (2)
Somewhat like me (3)
A little like me (4)
Not like me (5)
Not at all like me (6) 
\end{lstlisting}

To calculate the 0-1 value identification score $s_{\text{real}}$ of real individuals from question answer $q$, consider the meaning of answer numbers (4 and 5 is the borderline), we use $$s_{\text{real}}=\text{max}((4.5-q)/3.5, 0).$$ These questions (v70-v79) sequentially correspond to Schwartz values \textit{Self-direction, Power, Security, Hedonism, Benevolence, Achievement, Stimulation, Conformity, Universalism, Tradition}. Accordingly, we get the value identification answers on 10 values for real individuals.

For each individual, we input the content of these questions and their responses to these questions into GPT-4, and guided GPT-4 to mimic an \textit{virtual individual possessing these values} through prompts \cite{aher2023using}. Subsequently, we selected 20 social topics and guided the simulated individuals to express and explain their views on these social topics, forming 20 text data instances for each individual with format similar to Touché23-ValueEval. These topics are:

\begin{table*}[ht]
    \begin{adjustbox}{width=1.9\columnwidth,center}
    \begin{tabular}{l|c|c|c}
        \toprule
        Method & Inference Cost (3.4k samples) & Training Cost & Test Set F1 Score\\
        \midrule
        GPT-4o-mini (\textit{single-step} prompting) & \$1.22 & - & 0.53 \\
        GPT-4o-mini (\textit{sequential} prompting - CoT) & \$1.90 & - & 0.57  \\
        GPT-4-turbo (\textit{sequential} prompting - CoT) & \$37.9 & - & 0.58  \\
        \midrule
        EAVIT (Llama2-chat-13b + GPT-4o-mini) & \textbf{\$0.27} & \$3.9 (8k generations) & \textbf{0.69}  \\
        
        \bottomrule
    \end{tabular}
    \end{adjustbox}
    \caption{API cost evaluation on ValueEval.}
    \label{tbla1}
\end{table*}

\begin{table*}[ht]
    \begin{adjustbox}{width=0.7\columnwidth,center}
    \begin{tabular}{c|c|c}
        \toprule
        $p_\text{low}$ and $p_\text{high}$& Test Set F1 Score & \# LLM tokens\\
        \midrule
        $0.2, 0.8$ & \textbf{0.67} & 0.51k \\
        $0, 0.8$ & 0.59 & 2.45k \\
        $0.2, 1.0$ & 0.65 & 0.67k\\
        $0.4, 0.8$ & 0.62 & 0.38k \\
        
        \bottomrule
    \end{tabular}
    \end{adjustbox}
    \caption{Ablation study of candidate value set thresholds of EAVIT (Llama2-chat-13b + GPT-4o-mini) on ValueEval.}
    \label{tbla2}
\end{table*}

\begin{lstlisting}
"We should allocate 3% of GDP for research and innovation by 2035."
"We should abandon the use of school uniform."
"We should subsidize space exploration."
"We should abandon television."
"We should invest in the development of high-speed railways."
"We should limit executive compensation."
"College education should be free for everyone."
"We must respect our parents."
"We should protect our privacy in the Internet age."
"We should ban private military companies."
"We should carry on the habits of our elders."
"We should end illegal pushbacks."
"We often unconsciously conform to others' opinions."
"We should not be complacent."
"We should be kind to everyone."
"Retirement homes should become more common."
"Equal pay for equal work should apply as a principle."
"We should stop unrestricted deforestation and logging."
"Carbon emissions trading should be limited."
"We should adopt atheism."
\end{lstlisting}
The complete set of generated data is attached to the paper. A sample of the generated text data instances are:
\begin{lstlisting}

 {"argument": "We should allocate 3% of GDP for research and innovation by 2035.", "answer": "supportive", "explanation": "I believe it is important to allocate 3% of GDP for research and innovation because it allows for the development of new ideas and creativity, which I value. Additionally, investing in research and innovation can lead to advancements that benefit society and improve the well-being of people around me."}

 {"argument": "We should abandon television.", "answer": "unsupportive", "explanation": "I am unsupportive of the idea to abandon television because it is very important to me to have a good time and spoil myself. Television provides entertainment and relaxation, allowing me to unwind and enjoy myself after a long day."}

 {"argument": "College education should be free for everyone. ", "answer": "supportive", "explanation": "I believe that college education should be free for everyone. It is important to help society and provide equal opportunities for all. Education is a stepping stone for success and by making it accessible to all, we can create a more equal and prosperous society."}

 {"argument": "We should end illegal pushbacks. ", "answer": "supportive", "explanation": "I believe that we should end illegal pushbacks because it aligns with my values of doing something for the good of society and caring for the well-being of others. It is important to prioritize human rights and ensure fair and just treatment for everyone."}
\end{lstlisting}

Finally, we process these text data using EAVIT and other methods trained on Touché23-ValueEval to identify values behind them. To aggregate the results on each text data into the final value identification result for individuals, we label the result $s_{\text{EAVIT}}$ of each virtual individual on the value $v$ as 1 if and only if there are at least 3 associated generated texts that are relevant to the value $v$. In table \ref{tbl_casestudy}, we classify the result $s_{\text{EAVIT}}$ as accurate if $|s_{\text{EAVIT}} - s_{\text{real}}|<0.5$.

\subsubsection{Training Cost Evaluation}
\label{asec1}
Table \ref{tbla1} is an evaluation of OpenAI API cost on experiments on ValueEval. The inference API cost of EAVIT is signifcantly lower than complete LLM-based methods. Besides, the cost for generating additional training data is also lower than the entire inference API cost of GPT-4o-mini (\textit{single-step} prompting) method, showing that the data generation process is cost-effective for performance improvement. The local LM (Llama2-13b) is 4-bit quantized and trained on a single local RTX 4090 GPU for 10 hours, which is much lower than the online LLM costs. 

\subsubsection{Experiments on Thresholds $p_\text{low}$ and $p_\text{high}$}
\label{asec2}
Table \ref{tbla2} is an ablation study of the thresholds $p_\text{low}$ and $p_\text{high}$ in EAVIT, which control the selection of candidate values to LLM for final identification \ref{candidate_set}. According to the results, the current experimental set $p_\text{low}=0.2$ and $p_\text{high}=0.8$ is the optimal setting.

\subsection{More Discussion}
\paragraph{Human Value Theories }In psychology, research on human values has been extensive, with key contributions from \cite{rokeach1973nature} establishing a value system, \cite{schwartz2012overview} creating Schwartz's theory of basic values and cross-cultural value questions, \cite{cheng2010developing} consolidating taxonomies into a meta-inventory of values, \cite{gert2004common} proposing common morality theory, \cite{graham2013moral} building the Moral Foundation Theory, and \cite{de2022basic} introducing the linked open data scheme, ValueNet. These taxonomies have been extensively utilized in social science studies and automated detection of human values could significantly aid such analyses \cite{scharfbillig2021values,scharfbillig2022monitoring}. In our paper, following existing works \cite{kiesel2023semeval}, we adopt Schwartz's Theory of Basic Values \cite{schwartz2012overview} as the basic value system for value identification, which has been applied in multiple fields including economics \cite{ng2005predictors} and LLMs \cite{miotto2022gpt,fischer2023does}. However, it should be noted that our method is applicable to any completely defined value system (such as the extended Schwartz value system or those with values set for language models).

\end{document}